\documentclass[11pt,a4paper]{article}
\usepackage[hyperref]{emnlp2020}
\usepackage{times}
\usepackage{latexsym}

\usepackage{microtype}

\usepackage{amsfonts}
\usepackage{graphicx}
\usepackage{mathtools}
\usepackage{multirow}
\usepackage{paralist}

\usepackage{color}

\aclfinalcopy 
\def\aclpaperid{327} 

\title{Global-to-Local Neural Networks for Document-Level\\ Relation Extraction}

\author{Difeng Wang$^\dag$ \quad Wei Hu$^{\dag,\,\ddag,\,}$\thanks{\ \ Corresponding author} \quad Ermei Cao$^\dag$ \quad Weijian Sun$^\S$\\
	$^\dagger$ State Key Laboratory for Novel Software Technology, Nanjing University, China \\
	$^\ddagger$ National Institute of Healthcare Data Science, Nanjing University, China\\
	$^\S$ Huawei Technologies Co., Ltd. \\
	\texttt{\{dfwang,emcao\}.nju@gmail.com,\,whu@nju.edu.cn,\,sunweijian@huawei.com} 
}

\date{}

\begin{document}
\maketitle

\begin{abstract}
Relation extraction (RE) aims to identify the semantic relations between named entities in text. Recent years have witnessed it raised to the document level, which requires complex reasoning with entities and mentions throughout an entire document. In this paper, we propose a novel model to document-level RE, by encoding the document information in terms of entity global and local representations as well as context relation representations. Entity global representations model the semantic information of all entities in the document, entity local representations aggregate the contextual information of multiple mentions of specific entities, and context relation representations encode the topic information of other relations. Experimental results demonstrate that our model achieves superior performance on two public datasets for document-level RE. It is particularly effective in extracting relations between entities of long distance and having multiple mentions.
\end{abstract}

\section{Introduction}

Relation extraction (RE) aims to identify the semantic relations between named entities in text. While previous work \cite{zeng2014relation,zhang2015bidirectional,zhang2018graph} focuses on extracting relations within a sentence, a.k.a.~\emph{sentence}-level RE, recent studies \cite{verga2018simultaneously,christopoulou2019connecting,sahu2019inter,yao2019docred} have escalated it to the \emph{document} level, since a large amount of relations between entities usually span across multiple sentences in the real world. According to an analysis on Wikipedia corpus \cite{yao2019docred}, at least 40.7\% of relations can only be extracted on the document level.

Compared with sentence-level RE, document-level RE requires more complex reasoning, such as logical reasoning, coreference reasoning and common-sense reasoning. A document often contains many entities, and some entities have multiple mentions under the same phrase of alias. To identify the relations between entities appearing in different sentences, document-level RE models must be capable of modeling the complex interactions between multiple entities and synthesizing the context information of multiple mentions. 

\begin{figure}
	\centering
	\includegraphics[width=\columnwidth]{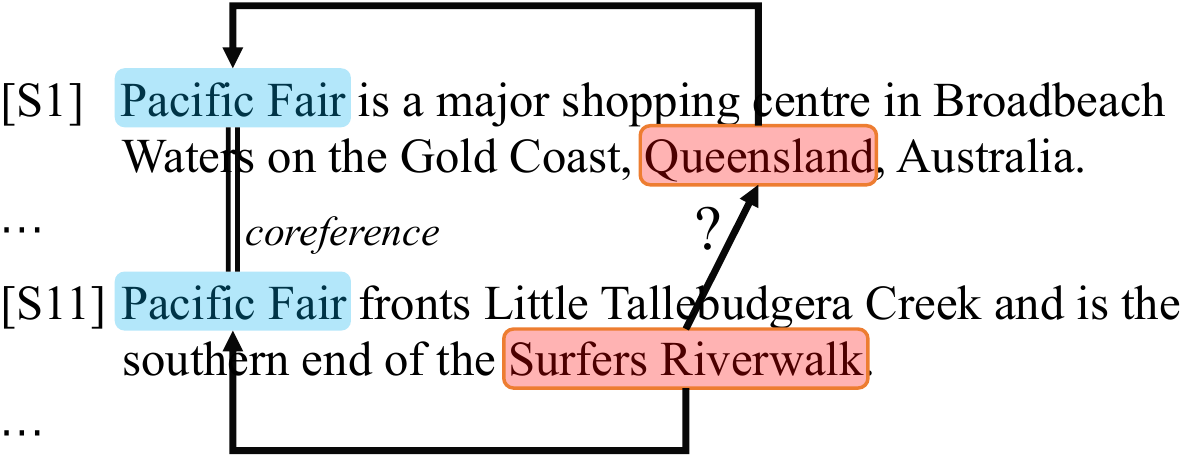}
	\caption{An example of document-level RE excerpted from the DocRED dataset~\cite{yao2019docred}. Arrows denote intra/inter-sentential relations.}
	\label{fig:example}
\end{figure}

Figure~\ref{fig:example} shows an example of document-level RE. Assume that one wants to extract the relation between \textit{``Surfers Riverwalk"} in S11 and \textit{``Queensland"} in S1. One has to find that \textit{``Surfers Riverwalk"} contains \textit{``Pacific Fair"} (from S11), and  \textit{``Pacific Fair"} (coreference) is located in \textit{``Queensland"} (from S1). This chain of interactions helps infer the inter-sentential relation \textit{``located in"} between \textit{``Surfers Riverwalk"} and \textit{``Queensland"}.

\smallskip
\noindent\textbf{State-of-the-art.} Early studies \cite{peng2017cross,quirk2017distant} confined document-level RE to short text spans (e.g., within three sentences). Some other studies \cite{nguyen2018convolutional,gupta2019neural} were restricted to handle two entity mentions in a document. We argue that they are incapable of dealing with the example in Figure~\ref{fig:example}, which needs to consider multiple mentions of entities integrally. To encode the semantic interactions of multiple entities in long distance, recent work defined document-level graphs and proposed graph-based neural network models. For example, \citet{sahu2019inter,gupta2019neural} interpreted words as nodes and constructed edges according to syntactic dependencies and sequential information. However, there is yet a big gap between word representations and relation prediction. \citet{christopoulou2019connecting} introduced the notion of document graphs with three types of nodes (mentions, entities and sentences), and proposed an edge-oriented graph neural model for RE. However, it indiscriminately integrated various information throughout the whole document, thus irrelevant information would be involved as noise and damages the prediction accuracy.

\smallskip
\noindent\textbf{Our approach and contributions.} To cope with the above limitations, we propose a novel graph-based neural network model for document-level RE. Our key idea is to make full use of document semantics and predict relations by learning the representations of involved entities from both coarse-grained and fine-grained perspectives as well as other context relations. Towards this goal, we address three challenges below:

First, \emph{how to model the complex semantics of a document?} We use the pre-trained language model BERT \cite{devlin2019bert} to capture semantic features and common-sense knowledge, and build a heterogeneous graph with heuristic rules to model the complex interactions between all mentions, entities and sentences in the document.

Second, \emph{how to learn entity representations effectively?} We design a global-to-local neural network to encode coarse-grained and fine-grained semantic information of entities. Specifically, we learn entity global representations by employing R-GCN \cite{schlichtkrull2018modeling} on the created heterogeneous graph, and entity local representations by aggregating multiple mentions of specific entities with multi-head attention \cite{vaswani2017attention}.

Third, \emph{how to leverage the influence from other relations?} In addition to target relation representations, other relations imply the topic information of a document. We learn context relation representations with self-attention \cite{sorokin2017context} to make final relation prediction.

In summary, our main contribution is twofold:
\begin{compactitem}
\item We propose a novel model, called \emph{GLRE}, for document-level RE. To predict relations between entities, GLRE synthesizes entity global representations, entity local representations and context relation representations integrally. For details, please see Section~\ref{sect:model}.
	
\item We conducted extensive experiments on two public document-level RE datasets. Our results demonstrated the superiority of GLRE compared with many state-of-the-art competitors. Our detailed analysis further showed its advantage in extracting relations between entities of long distance and having multiple mentions. For details, please see Section~\ref{sect:exp}.
\end{compactitem}


\begin{figure*}
	\centering
	\includegraphics[width=\textwidth]{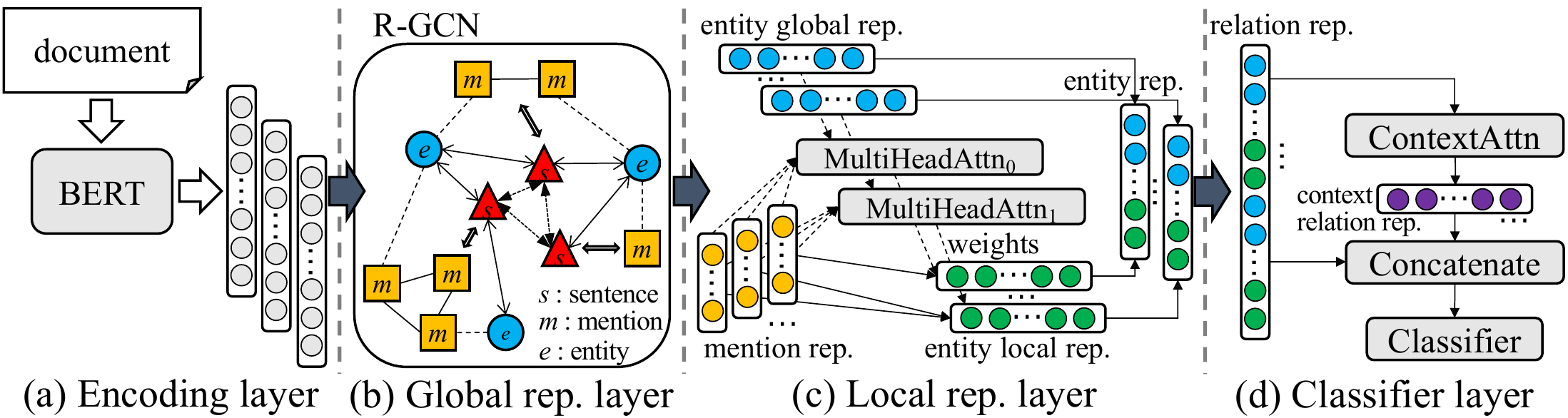}
	\caption{Architecture of the proposed model.}
	\label{fig:model}
\end{figure*}

\section{Related Work}
\label{sect:work}

RE has been intensively studied in a long history. 
In this section, we review closely-related work.

\smallskip
\noindent\textbf{Sentence-level RE.} Conventional work addressed sentence-level RE by using carefully-designed patterns \cite{soderland1995crystal}, features \cite{kambhatla2004combining} and kernels \cite{culotta2004dependency}. Recently, deep learning-based work has advanced the state-of-the-art without heavy feature engineering. Various neural networks have been exploited, e.g., CNN \cite{zeng2014relation}, RNN \cite{zhang2015bidirectional,cai2016bidirectional} and GNN \cite{zhang2018graph}. Furthermore, to cope with the wrong labeling problem caused by distant supervision, \citet{zeng2015distant} adopted Piecewise CNN (PCNN), \citet{lin2016neural,zhang2017position} employed attention mechanisms, and \citet{zhang2019long,qu2019a} leveraged knowledge graphs as external resources. All these models are limited to extracting intra-sentential relations. They also ignore the interactions of entities outside a target entity pair.

\smallskip
\noindent\textbf{Document-level RE.} As documents often provide richer information than sentences, there has been an increasing interest in document-level RE. \citet{gu2017chemical,nguyen2018convolutional,gupta2019neural,wang2019fine} extended the sentence-level RE models to the document level. \citet{ye2020coreferential} explicitly incorporated coreference information into language representation models (e.g., BERT).  \citet{zheng2018effective,tang2020hin} proposed hierarchical networks to aggregate information from the word, sentence and document levels.

\citet{quirk2017distant} proposed the notion of document-level graphs, where nodes denote words and edges incorporate both syntactic dependencies and discourse relations. Following this, \citet{peng2017cross} first splitted a document-level graph into two directed acyclic graphs (DAGs), then used a graph LSTM for each DAG to learn the contextual representation of each word, which was concatenated and finally fed to the relation classifier. Differently, \citet{song2018nary} kept the original graph structure and directly modeled the whole document-level graph using graph-state LSTM. These models only predict the relation of a single mention pair in a document at a time, and ignore multiple mentions of a target entity pair as well as other entities.

Several models predict the relation of a target entity pair by aggregating the scores of all mention pairs with multi-instance learning. \citet{verga2018simultaneously} proposed a Transformer-based model. Later, \citet{sahu2019inter} switched Transformer to GCN. The two models only consider one target entity pair per document, and construct the document-level graphs relying on external syntactic analysis tools. \citet{christopoulou2019connecting} built a document graph with heterogeneous types of nodes and edges, and proposed an edge-oriented model to obtain global representations for relation classification. Our model differs in further learning entity local representations to reduce the influence of irrelevant information and considering other relations in the document to refine the prediction. Recently, \citet{nan2020lsr} defined a document graph as a latent variable and induced it based on the structured attention. Unlike our work, it improves the performance of document-level RE models by optimizing the structure of the document graph.

Besides, a few models \cite{levy2017zero,qiu2018qa4ie} borrowed the reading comprehension techniques to document-level RE. However, they require domain knowledge to design question templates, and may perform poorly in zero-answer and multi-answers scenarios \cite{liu2019neural}, which are very common for RE.

\section{Proposed Model}
\label{sect:model}

We model document-level RE as a \emph{classification} problem. Given a document annotated with entities and their corresponding textual mentions, the objective of document-level RE is to identify the relations of all entity pairs in the document.

Figure~\ref{fig:model} depicts the architecture of our model, named GLRE. It receives an entire document with annotations as input. First, in (a) \emph{encoding layer}, it uses a pre-trained language model such as BERT \cite{devlin2019bert} to encode the document. Then, in (b) \emph{global representation layer}, it constructs a global heterogeneous graph with different types of nodes and edges, and encodes the graph using a stacked R-GCN \cite{schlichtkrull2018modeling} to capture entity global representations. Next, in (c) \emph{local representation layer}, it aggregates multiple mentions of specific entities using multi-head attention \cite{vaswani2017attention} to obtain entity local representations. Finally, in (d) \emph{classifier layer}, it combines the context relation representations obtained with self-attention \cite{sorokin2017context} to make final relation prediction. Please see the rest of this section for technical details.

\subsection{Encoding Layer}

Let $\mathcal{D}=[w_1,w_2,\ldots,w_k]$ be an input document, where $w_j$ ($1\leq j\leq k$) is the $j^\textrm{th}$ word in it. We use BERT to encode $\mathcal{D}$ as follows:
\begin{align}
\resizebox{.89\columnwidth}{!}{$\begin{aligned}
\mathbf{H} = [\mathbf{h}_1,\mathbf{h}_2,\ldots,\mathbf{h}_k] = \textrm{BERT}([w_1,w_2,\ldots,w_k]),
\end{aligned}$}
\end{align}
where $\mathbf{h}_j\in\mathbb{R}^{d_w}$ is a sequence of hidden states at the output of the last layer of BERT. Limited by the input length of  BERT, we encode a long document sequentially in form of short paragraphs.

\subsection{Global Representation Layer}

Based on $\mathbf{H}$, we construct a \emph{global heterogeneous graph}, with different types of nodes and edges to capture different dependencies (e.g., co-occurrence dependencies, coreference dependencies and order dependencies), inspired by \citet{christopoulou2019connecting}. 
Specifically, there are three types of nodes:
\begin{compactitem}
\item \emph{Mention nodes,} which model different mentions of entities in $\mathcal{D}$. The representation of a mention node $m_i$ is defined by averaging the representations of contained words. To distinguish node types, we concatenate a node type representation $\mathbf{t}_m\in\mathbb{R}^{d_t}$. Thus, the representation of $m_i$ is $\mathbf{n}_{m_i} = [\mathrm{avg}_{w_j\in m_i} (\mathbf{h}_j); \mathbf{t}_m]$, where $[\,;]$ is the concatenation operator.

\item \emph{Entity nodes,} which represent entities in $\mathcal{D}$. The representation of an entity node $e_i$ is defined by averaging the representations of the mention nodes to which they refer, together with a node type representation $\mathbf{t}_e\in\mathbb{R}^{d_t}$. Therefore, the representation of $e_i$ is $\mathbf{n}_{e_i} = [\mathrm{avg}_{m_j\in e_i} (\mathbf{n}_{m_j}); \mathbf{t}_e]$.

\item \emph{Sentence nodes,} which encode sentences in $\mathcal{D}$. Similar to mention nodes, the representation of a sentence node $s_i$ is formalized as $\mathbf{n}_{s_i} = [\mathrm{avg}_{w_j\in s_i} (\mathbf{h}_j); \mathbf{t}_s]$, where $\mathbf{t}_s\in\mathbb{R}^{d_t}$.
\end{compactitem}

Then, we define five types of edges to model the interactions between the nodes:
\begin{compactitem}
\item \emph{Mention-mention edges.} We add an edge for any two mention nodes in the same sentence.

\item \emph{Mention-entity edges.} We add an edge between a mention node and an entity node if the mention refers to the entity.

\item \emph{Mention-sentence edges.} We add an edge between a mention node and a sentence node if the mention appears in the sentence.

\item \emph{Entity-sentence edges.} We create an edge between an entity node and a sentence node if at least one mention of the entity appears in the sentence.

\item \emph{Sentence-sentence edges.} We connect all sentence nodes to model the non-sequential information (i.e., break the sentence order).
\end{compactitem}

Note that there are no entity-entity edges, because they form the relations to be predicted.

Finally, we employ an $L$-layer stacked R-GCN \cite{schlichtkrull2018modeling} to convolute the global heterogeneous graph. Different from GCN, R-GCN considers various types of edges and can better model multi-relational graphs. Specifically, its node forward-pass update for the $(l+1)^\textrm{th}$ layer is defined as follows:
\begin{align}
\resizebox{.89\columnwidth}{!}{$\begin{aligned}
\mathbf{n}^{l+1}_i = \sigma\Big( \sum_{x \in\mathcal{X}} \sum_{j \in\mathcal{N}^x_i} \frac{1}{|\mathcal{N}_i^x|} \mathbf{W}^l_x\mathbf{n}^l_j + \mathbf{W}^l_0\mathbf{n}^l_i \Big),
\end{aligned}$}
\end{align} 
where $\sigma(\cdot)$ is the activation function. $\mathcal{N}^x_i$ denotes the set of neighbors of node $i$ linked with edge $x$, and $\mathcal{X}$ denotes the set of edge types. $ \mathbf{W}^l_x,\mathbf{W}^l_0\in\mathbb{R}^{d_n \times d_n}$ are trainable parameter matrices ($d_n$ is the dimension of node representations).

We refer to the representations of entity nodes after graph convolution as \emph{entity global representations}, which encode the semantic information of entities throughout the whole document. We denote an entity global representation by $\mathbf{e}_i^\textrm{glo}$.

\subsection{Local Representation Layer}

We learn \emph{entity local representations} for specific entity pairs by aggregating the associated mention representations with multi-head attention  \cite{vaswani2017attention}. The ``local'' can be understood from two angles: (i) It aggregates the original mention information from the encoding layer. (ii) For different entity pairs, each entity would have multiple different local representations w.r.t. the counterpart entity. However, there is only one entity global representation.

Multi-head attention enables a RE model to jointly attend to the information of an entity composed of multiple mentions from different representation subspaces. Its calculation involves the sets of queries $\mathcal{Q}$ and key-value pairs $(\mathcal{K},\mathcal{V})$:
\begin{align}
\resizebox{.89\columnwidth}{!}{$\begin{aligned}
\mathrm{MHead}(\mathcal{Q},\mathcal{K},\mathcal{V}) = [\textrm{head}_1;\ldots;\textrm{head}_z] \mathbf{W}^\textrm{out},
\end{aligned}$} \\
\resizebox{.89\columnwidth}{!}{$\begin{aligned}
\textrm{head}_i = \mathrm{softmax}\Big( \frac{\mathcal{Q} \mathbf{W}^\mathcal{Q}_i {(\mathcal{K} \mathbf{W}^\mathcal{K}_i)}^\prime} {\sqrt{d_v}} \Big) \mathcal{V} \mathbf{W}^\mathcal{V}_i,
\end{aligned}$}
\end{align}
where $\mathbf{W}^\textrm{out}\in\mathbb{R}^{d_n \times d_n}$ and $\mathbf{W}^\mathcal{Q}_i, \mathbf{W}^\mathcal{K}_i, \mathbf{W}^\mathcal{V}_i\in\mathbb{R}^{d_n\times d_v}$ are trainable parameter matrices. $z$ is the number of heads satisfying that $z\times d_v = d_n$.

In this paper, $\mathcal{Q}$ is related to the entity global representations, $\mathcal{K}$ is related to the initial sentence node representations before graph convolution (i.e., the input features of sentence nodes in R-GCN), and $\mathcal{V}$ is related to the initial mention node representations. 
Specifically, given an entity pair $(e_a,e_b)$, we define their local representations as follows:
\begin{align}
\resizebox{\columnwidth}{!}{$\begin{aligned}
\mathbf{e}_a^\textrm{loc} &= \mathrm{LN}\big( \mathrm{MHead}_0(\mathbf{e}_b^\textrm{glo}, \{\mathbf{n}_{s_i}\}_{s_i\in \mathcal{S}_a}, \{\mathbf{n}_{m_j}\}_{m_j\in \mathcal{M}_a}) \big), \\
\mathbf{e}_b^\textrm{loc} &= \mathrm{LN}\big( \mathrm{MHead}_1(\mathbf{e}_a^\textrm{glo}, \{\mathbf{n}_{s_i}\}_{s_i\in \mathcal{S}_b}, \{\mathbf{n}_{m_j}\}_{m_j\in \mathcal{M}_b}) \big), \\
\end{aligned}$}
\end{align} 
where $\mathrm{LN}(\cdot)$ denotes layer normalization \cite{ba2016layer}. $\mathcal{M}_a$ is the corresponding mention node set of $e_a$, and $\mathcal{S}_a$ is the corresponding sentence node set in which each mention node in $\mathcal{M}_a$ is located. $\mathcal{M}_b$ and $\mathcal{S}_b$ are similarly defined for $e_b$. Note that $\mathrm{MHead}_0$ and $\mathrm{MHead}_1$ learn independent model parameters for entity local representations.

Intuitively, if a sentence contains two mentions $m_a,m_b$ corresponding to $e_a,e_b$, respectively, then the mention node representations $\mathbf{n}_{m_a},\mathbf{n}_{m_b}$ should contribute more to predicting the relation of $(e_a,e_b)$ and the attention weights should be greater in getting $\mathbf{e}_a^\textrm{loc},\mathbf{e}_b^\textrm{loc}$. More generally, a higher semantic similarity between the node representation of a sentence containing $m_a$ and $\mathbf{e}^\textrm{glo}_b$ indicates that this sentence and $m_b$ are more semantically related, and $\textbf{n}_{m_a}$ should get a higher attention weight to $\mathbf{e}^\textrm{loc}_a$.

\subsection{Classifier Layer}

To classify the target relation $r$ for an entity pair $(e_a,e_b)$, we firstly concatenate entity global representations, entity local representations and relative distance representations to generate entity final representations: 
\begin{align}
\begin{aligned}
\mathbf{\hat{e}}_a &= [\mathbf{e}_a^\textrm{glo}; \mathbf{e}_a^\textrm{loc}; \mathbf{\Delta}(\delta_{ab})], \\
\mathbf{\hat{e}}_b &= [\mathbf{e}_b^\textrm{glo}; \mathbf{e}_b^\textrm{loc}; \mathbf{\Delta}(\delta_{ba})],
\end{aligned}
\end{align}
where $\delta_{ab}$ denotes the relative distance from the first mention of $e_a$ to that of $e_b$ in the document. $\delta_{ba}$ is similarly defined. The relative distance is first divided into several bins $\{1,2,\ldots,2^b\}$. Then, each bin is associated with a trainable distance embedding. $\mathbf{\Delta}(\cdot)$ associates each $\delta$ to a bin.

Then, we concatenate the final representations of $e_a,e_b$ to form the \emph{target relation representation} $\mathbf{o}_r = [\mathbf{\hat{e}}_a; \mathbf{\hat{e}}_b]$.

Furthermore, all relations in a document implicitly indicate the topic information of the document, such as \textit{``director"} and \textit{``character"} often appear in movies. In turn, the topic information implies possible relations. Some relations under similar topics are likely to co-occur, while others under different topics are not. Thus, we use self-attention  \cite{sorokin2017context} to capture \emph{context relation representations}, which reveal the topic information of the document:
\begin{align}
\resizebox{.89\columnwidth}{!}{$\begin{aligned}
\mathbf{o}_c = \sum_{i=0}^p \theta_i \mathbf{o}_i = \sum_{i=0}^p \frac{\mathrm{exp}(\mathbf{o}_i \mathbf{W} \mathbf{o}_r^\prime)}
{\sum_{j=0}^p \mathrm{exp}(\mathbf{o}_j \mathbf{W} \mathbf{o}_r^\prime)} \mathbf{o}_i,
\end{aligned}$}
\end{align}
where $\mathbf{W}\in\mathbb{R}^{d_r\times d_r}$ is a trainable parameter matrix. $d_r$ is the dimension of target relation representations. $\mathbf{o}_i$ ($\mathbf{o}_j$) is the relation representation of the $i^\textrm{th}$ ($j^\textrm{th}$) entity pair. $\theta_i$ is the attention weight for $\mathbf{o}_i$. $p$ is the number of entity pairs.

Finally, we use a feed-forward neural network (FFNN) over the target relation representation $\mathbf{o}_r$ and the context relation representation $\mathbf{o}_c$ to make the prediction. Besides, considering that an entity pair may hold several relations, we transform the multi-classification problem into multiple binary classification problems. The predicted probability distribution of $r$ over the set $\mathcal{R}$ of all relations is defined as follows:
\begin{align}
\mathbf{y}_r = \mathrm{sigmoid}(\mathrm{FFNN}([\mathbf{o}_r; \mathbf{o}_c])),
\end{align}
where $\mathbf{y}_r\in\mathbb{R}^{|\mathcal{R}|}$.

We define the loss function as follows:
\begin{align}
\resizebox{.89\columnwidth}{!}{$\begin{aligned}
\mathcal{L} = -\sum_{r\in \mathcal{R}} \Big( y_r^* \log(y_r) + (1-y_r^*) \log(1-y_r) \Big),
\end{aligned}$}
\end{align}
where $y_r^*\in\{0,1\}$ denotes the true label of $r$. We employ Adam optimizer \cite{kingma2015adam} to optimize this loss function.
	
\section{Experiments and Results}
\label{sect:exp}

We implemented our GLRE with PyTorch 1.5. The source code and datasets are available online.\footnote{\url{https://github.com/nju-websoft/GLRE}} In this section, we report our experimental results.

\subsection{Datasets}

We evaluated GLRE on two public document-level RE datasets. Table \ref{tab:stat} lists their statistical data:
\begin{compactitem} 
\item The Chemical-Disease Relations (\emph{CDR}) data set \cite{li2016biocreative} was built for the BioCreative V challenge and annotated with one relation \textit{``chemical-induced disease''} manually.
	
\item The \emph{DocRED} dataset \cite{yao2019docred} was built from Wikipedia and Wikidata, covering various relations related to science, art, personal life, etc. Both manually-annotated and distantly-supervised data are offered. We only used the manually-annotated data.
\end{compactitem}

\begin{table}
\centering
\resizebox{\columnwidth}{!}{
	\begin{tabular}{|l|l|r|r|r|r|}
		\hline \multicolumn{2}{|c|}{Datasets} & \#Doc. & \#Rel. & \#Inst. & \#N/A Inst. \\ 
		\hline \multirow{3}{*}{CDR} & Train & 500 & 1 & 1,038 & 4,280 \\
			~ & Dev. & 500 & 1 & 1,012 & 4,136 \\
			~ & Test & 500 & 1 & 1,066 & 4,270 \\
		\hline \multirow{3}{*}{DocRED} & Train & 3,053 & 96 & 38,269 & 1,163,035 \\
			~ & Dev. & 1,000 & 96 & 12,332 & 385,263 \\
			~ & Test & 1,000 & 96 & 12,842 & 379,316 \\
		\hline
	\end{tabular}}
\caption{Dataset statistics (Inst.: relation instances excluding N/A relation; N/A Inst.: negative examples).}
\label{tab:stat} 
\end{table}

\subsection{Comparative Models}

First, we compared GLRE with five sentence-level RE models adapted to the document level:
\begin{compactitem}
\item \citet{zhang2018graph} employed GCN over pruned dependency trees.
\item \citet{yao2019docred} proposed four baseline models. The first three ones are based on CNN, LSTM and BiLSTM, respectively. The fourth context-aware model incorporates the attention mechanism into LSTM.
\end{compactitem}

We also compared GLRE with nine document-level RE models: 
\begin{compactitem}
\item \citet{zhou2016exploiting} combined feature-, tree kernel- and neural network-based models.
\item \citet{gu2017chemical} leveraged CNN and maximum entropy.
\item \citet{nguyen2018convolutional} integrated character-based word representations in CNN.
\item \citet{panyam2018exploiting} exploited graph kernels.
\item \citet{verga2018simultaneously} proposed a bi-affine network with Transformer.
\item \citet{zheng2018effective} designed a hierarchical network using multiple BiLSTMs.
\item \citet{christopoulou2019connecting} put forward an edge-oriented graph neural model with multi-instance learning.
\item \citet{wang2019fine} applied BERT to encode documents, and used a bilinear layer to predict entity relations. It improved performance by two phases. First, it predicted whether a relation exists between two entities. Then, it predicted the type of the relation.
\item \citet{tang2020hin} is a sequence-based model. It also leveraged BERT and designed a hierarchical inference network to aggregate inference information from entity level to sentence level, then to document level.
\end{compactitem}

\subsection{Experiment Setup}

Due to the small size of CDR, some work \cite{zhou2016exploiting,verga2018simultaneously,zheng2018effective,christopoulou2019connecting} created a new split by unionizing the training and development sets, denoted by \emph{``train + dev"}. Under this setting, a model was trained on the train + dev set, while the best epoch was found on the development set. To make a comprehensive comparison, we also measured the corresponding precision, recall and F1 scores.

For consistency, we used the same experiment setting on DocRED. Additionally, the gold standard of the test set of DocRED is unknown, and only F1 scores can be obtained via an online interface. Besides, it was noted that some relation instances are present in both training and development/test sets \cite{yao2019docred}. We also measured F1 scores ignoring those duplicates, denoted by \emph{Ign F1}.

For GLRE and \citet{wang2019fine}, we used different BERT models in the experiments. For CDR, we chose BioBERT-Base v1.1 \cite{lee2019biobert}, which re-trained the BERT-Base-cased model on biomedical corpora. For DocRED, we picked up the BERT-Base-uncased model. For the comparative models without using BERT, we selected the PubMed pre-trained word embeddings \cite{chiu2016train} for CDR and GloVe \cite{pennington2014glove} for DocRED. For the models with source code, we used our best efforts to tune the hyperparameters. Limited by the space, we refer interested readers to the appendix for more details.

\subsection{Main Results}

\begin{table}
\centering
\resizebox{\columnwidth}{!}{
	\begin{tabular}{l|ccc|ccc}
		\hline \multirow{2}{*}{\textbf{Models}} & \multicolumn{3}{c|}{\textbf{Train}} & \multicolumn{3}{c}{\textbf{Train\,+\,Dev}} \\ 
		\cline{2-7} & P & R & F1 & P & R & F1 \\
		\hline 	\citeauthor{zhang2018graph}$^\P$ & 52.3 & \underline{72.0} & 60.6 & 58.1 & \textbf{74.6} & 65.3 \\
		\hline	
				\citeauthor{zhou2016exploiting} & \underline{64.9} & 49.3 & 56.0 & 55.6 & 68.4 & 61.3 \\
				\citeauthor{gu2017chemical} & 55.7 & 68.1 & 61.3 & - & - & - \\
				\citeauthor{nguyen2018convolutional} & 57.0 & 68.6 & 62.3 & - & - & - \\
				\citeauthor{panyam2018exploiting} & 55.6 & 68.4 & 61.3 & - & - & - \\
				\citeauthor{verga2018simultaneously} & 55.6 & 70.8 & 62.1 & 63.3 & 67.1 & 65.1 \\
				\citeauthor{zheng2018effective} & 45.2 & 68.1 & 54.3 & 56.2 & 68.0 & 61.5 \\
				\citeauthor{christopoulou2019connecting}$^\P$ & 62.7 & 66.3 & 64.5 & 61.5 & 73.6 & 67.0 \\
				\citeauthor{wang2019fine}$^\P$  & 61.9 & 68.7 & \underline{65.1} & \underline{66.0} & 68.3 & \underline{67.1} \\
		\hline	GLRE (ours) & \textbf{65.1} & \textbf{72.2} & \textbf{68.5} & \textbf{70.5} & \underline{74.5} & \textbf{72.5} \\
		\hline 	\multicolumn{7}{l}{$^\P$ denotes that we performed hyperparameter tuning. For others,} \\
				\multicolumn{7}{l}{\ \ \ we reused the reported results due to the lack of source code.}
	\end{tabular}}
\caption{Result comparison on CDR.}
\label{tab:cdr}
\end{table}

\begin{table}
\centering
\resizebox{\columnwidth}{!}{
	\begin{tabular}{l|cc|cc}
		\hline \multirow{2}{*}{\textbf{Models}} & \multicolumn{2}{c|}{\textbf{Train}} & \multicolumn{2}{c}{\textbf{Train\,+\,Dev}} \\ 
		\cline{2-5} & Ign F1 & F1 & Ign F1 & F1 \\
		\hline 	\citeauthor{zhang2018graph}$^\P$ & 49.9 & 52.1 & 52.5 & 54.6 \\
				\citeauthor{yao2019docred} (CNN) & 40.3 & 42.3 & - & - \\
				\citeauthor{yao2019docred} (LSTM) & 47.7 & 50.1 & - & - \\
				\citeauthor{yao2019docred} (BiLSTM) & 48.8 & 51.1 & - & - \\
				\citeauthor{yao2019docred} (Context-aware) & 48.4 & 50.7 & - & - \\
		\hline	\citeauthor{christopoulou2019connecting}$^\P$ & 49.1 & 50.9 & 48.3 & 50.4 \\
				\citeauthor{wang2019fine}$^\P$ & 53.1 & 55.4 & \underline{54.5} & \underline{56.5} \\
				\citeauthor{tang2020hin} & \underline{53.7} & \underline{55.6} & - & - \\
		\hline	GLRE (ours) & \textbf{55.4} & \textbf{57.4} & \textbf{56.7} & \textbf{58.9} \\
		\hline
	\end{tabular}}
\caption{Result comparison on DocRED.}
\label{tab:docred}
\end{table}

Tables~\ref{tab:cdr} and \ref{tab:docred} list the results of the comparative models and GLRE on CDR and DocRED, respectively. We have four findings below:
\begin{compactenum}[(1)]
\item The sentence-level RE models \cite{zhang2018graph,yao2019docred} obtained medium performance. They still fell behind a few docu-ment-level models, indicating the difficulty of directly applying them to the document level.

\item The graph-based RE models \cite{panyam2018exploiting,verga2018simultaneously,christopoulou2019connecting} and the non-graph models \cite{zhou2016exploiting,gu2017chemical,nguyen2018convolutional,zheng2018effective} achieved comparable results, while the best graph-based model \cite{christopoulou2019connecting} outperformed the best non-graph \cite{nguyen2018convolutional}. We attribute it to the document graph on the entity level, which can better model the semantic information in a document.

\item From the results of \citet{wang2019fine,tang2020hin}, the BERT-based models showed stronger prediction power for document-level RE. They outperformed the other comparative models on both CDR and DocRED.

\item GLRE achieved the best results among all the models. We owe it to entity global and local representations. Furthermore, BERT and context relation representations also boosted the performance. See our analysis below.
\end{compactenum}

\begin{figure}
	\centering
	\includegraphics[width=\columnwidth]{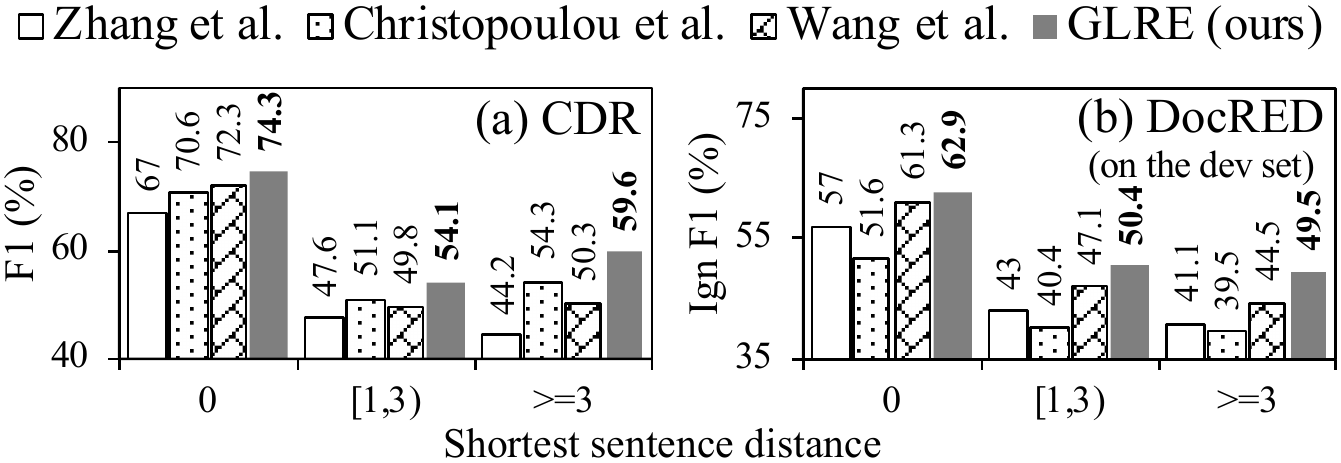}
	\caption{Results w.r.t. entity distance.}
	\label{fig:dist}
\end{figure}

\subsection{Detailed Analysis}

\noindent\textbf{Entity distance.} 
We examined the performance of the open-source models in terms of entity distance, which is defined as the shortest sentence distance between all mentions of two entities. Figure~\ref{fig:dist} depicts the comparison results on CDR and DocRED using the training set only. We observe that:
\begin{compactenum}[(1)]
\item GLRE achieved significant improvement in extracting the relations between entities of long distance, especially when $\textrm{distance}\geq 3$. This is because the global heterogeneous graph can effectively model the interactions of semantic information of different nodes (i.e., mentions, entities and sentences) in a document. Furthermore, entity local representations can reduce the influence of noisy context of multiple mentions of entities in long distance.

\item According to the results on CDR, the graph-based model \cite{christopoulou2019connecting} performed better than the sentence-level model \cite{zhang2018graph} and the BERT-based model \cite{wang2019fine} in extracting inter-sentential relations. The main reason is that it leveraged heuristic rules to construct the document graph at the entity level, which can better model the semantic information across sentences and avoid error accumulation involved by NLP tools, e.g., the dependency parser used in \citet{zhang2018graph}.

\item On DocRED, the models \cite{wang2019fine,zhang2018graph} outperformed the model \cite{christopoulou2019connecting}, due to the power of BERT and the increasing accuracy of dependency parsing in the general domain.
\end{compactenum}

\begin{figure}
	\centering
	\includegraphics[width=\columnwidth]{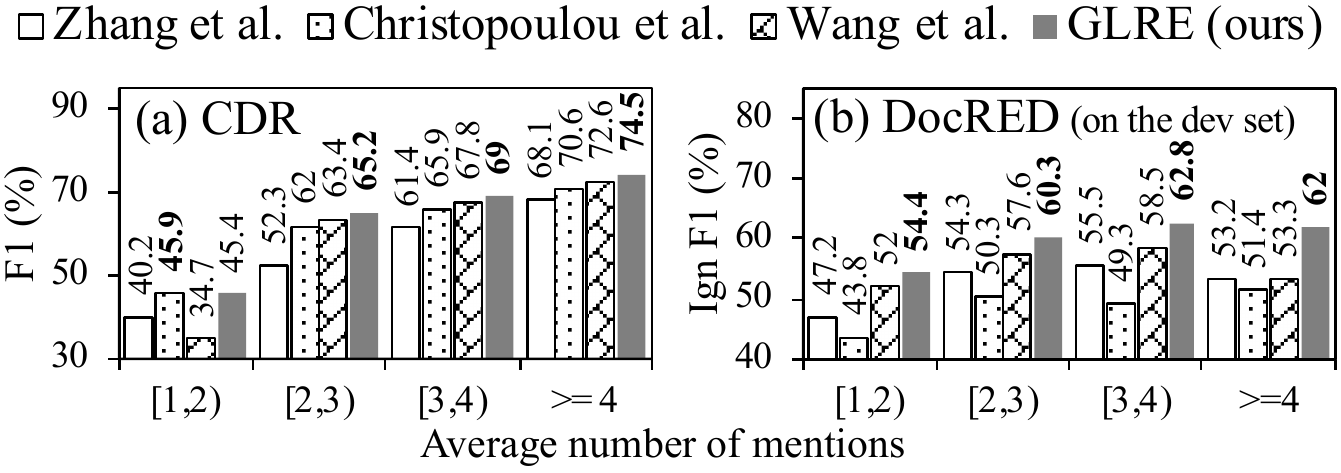}
	\caption{Results w.r.t. number of entity mentions.}
	\label{fig:mention}
\end{figure}

\smallskip
\noindent\textbf{Number of entity mentions.} To assess the effectiveness of GLRE in aggregating the information of multiple entity mentions, we measured the performance in terms of the average number of mentions for each entity pair. Similar to the previous analysis, Figure~\ref{fig:mention} shows the results on CDR and DocRED using the training set only. We see that:
\begin{compactenum}[(1)]
\item GLRE achieved great improvement in extracting the relations with average number of mentions $\geq$ 2, especially $\geq$ 4. The major reason is that  entity local representations aggregate the contextual information of multiple mentions selectively. As an exception, when the average number of mentions was in $[1,2)$, the performance of GLRE was slightly lower than \citet{christopoulou2019connecting} on CDR. This is because both GLRE and \citet{christopoulou2019connecting} relied on modeling the interactions between entities in the document, which made them indistinguishable under this case. In fact, the performance of all the models decreased when the average number of mentions was small, because less relevant information was provided in the document, which made relations harder to be predicted. We will consider external knowledge in our future work.

\item As compared with \citet{zhang2018graph} and \citet{christopoulou2019connecting}, the BERT-based model \cite{wang2019fine} performed better in general, except for one interval. When the average number of mentions was in $[1,2)$ on CDR, its performance was significantly lower than other models. The reason is twofold. On one hand, it is more difficult to capture the latent knowledge in the biomedical field. On the other hand, the model \cite{wang2019fine} only relied on the semantic information of the mentions of target entity pairs to predict the relations. When the average number was small, the prediction became more difficult. Furthermore, when the average number was large, its performance increase was not significant. The main reason is that, although BERT brought rich knowledge, the model \cite{wang2019fine} indiscriminately aggregated the information of multiple mentions and introduced much noisy context, which limited its performance. 
\end{compactenum}

\begin{table}
\centering
\resizebox{\columnwidth}{!}{
	\begin{tabular}{l|ccc|cc}
		\hline \multirow{2}{*}{\textbf{Models}} & \multicolumn{3}{c|}{\textbf{CDR}} & \multicolumn{2}{c}{\textbf{DocRED}} \\ 
		\cline{2-6} & P & R & F1 & Ign F1 & F1 \\
		\hline 	GLRE  & 65.1 & 72.2 & \textbf{68.5} & \textbf{55.4} & \textbf{57.4} \\
				\ \ w/o BERT & \textbf{69.6} & 66.5 & 68.0 & 51.6 & 53.6 \\
				\ \ w/o Entity global rep. & 67.0 & 65.4 & 66.2 & 54.7 & 56.6 \\
				\ \ w/o Entity local rep. & 60.9 & 68.5 & 64.5 & 54.6 & 56.4 \\
				\ \ w/o Context rel. rep. & 60.5 & \textbf{75.1} & 67.1 & 54.6 & 56.8 \\
		\hline
	\end{tabular}}
\caption{Results of ablation study.}
\label{tab:ablation}
\end{table}

\smallskip
\noindent\textbf{Ablation study.} To investigate the effectiveness of each layer in GLRE, we conducted an ablation study using the training set only. Table~\ref{tab:ablation} shows the comparison results. We find that: (1) BERT had a greater influence on DocRED than CDR. This is mainly because BERT introduced valuable linguistic knowledge and common-sense knowledge to RE, but it was hard to capture latent knowledge in the biomedical field. (2) F1 scores dropped when we removed entity global representations, entity local representations or context relation representations, which verified their usefulness in document-level RE. (3) Particularly, when we removed entity local representations, F1 scores dropped more dramatically. We found that more than 54\% and 19\% of entities on CDR and DocRED, respectively, have multiple mentions in different sentences. The local representation layer, which uses multi-head attention to selectively aggregate multiple mentions, can reduce much noisy context.

\smallskip
\noindent\textbf{Pre-trained language models.} To analyze the impacts of pre-trained language models on GLRE and also its performance upper bound, we replaced BERT-Base with BERT-Large, XLNet-Large \cite{yang2019xlnet} or ALBERT-xxLarge \cite{lan2020albert}. Table~\ref{tab:berts} shows the comparison results using the training set only, from which we observe that larger models boosted the performance of GLRE to some extent. When the ``train\,+\,dev" setting was used on DocRED, the Ign F1 and F1 scores of XLNet-Large even reached to 58.5 and 60.5, respectively. However, due to the lack of biomedical versions, XLNet-Large and ALBERT-xxLarge did not bring improvement on CDR. We argue that selecting the best pre-trained models is not our primary goal.

\begin{table}
\centering
\resizebox{\columnwidth}{!}{
	\begin{tabular}{l|ccc|cc}
		\hline \multirow{2}{*}{\textbf{GLRE}} & \multicolumn{3}{c|}{\textbf{CDR}} & \multicolumn{2}{c}{\textbf{DocRED}} \\ 
		\cline{2-6} & P & R & F1 & Ign F1 & F1 \\
		\hline 	BERT-Base & 65.1 & 72.2 & 68.5 & 55.4 & 57.4 \\
				BERT-Large & 65.3 & 72.3 & \textbf{68.6} & \textbf{56.8} & 58.9 \\
				XLNet-Large & \textbf{66.1} & 70.5 & 68.2 & \textbf{56.8} & \textbf{59.0} \\
				ALBERT-xxLarge & 57.5 & \textbf{80.6} & 67.1 & 56.3 & 58.3 \\
		\hline
	\end{tabular}}
\caption{Results w.r.t. different pre-training models.}
\label{tab:berts}
\end{table}

\smallskip
\noindent\textbf{Case study.} To help understanding, we list a few examples from the CDR test set in Table \ref{tab:case}. See Appendix for more cases from DocRED.
\begin{compactenum}[(1)]
\item From Case 1, we find that logical reasoning is necessary. Predicting the relation between \textit{``rofecoxib"} and \textit{``GI bleeding"} depends on the bridge entity \textit{``non-users of aspirin"}. GLRE used R-GCN to model the document information based on the global heterogeneous graph, thus it dealt with complex inter-sentential reasoning better.

\item  From Case 2, we observe that, when a sentence contained multiple entities connected by conjunctions (such as \textit{``and"}), the model \cite{wang2019fine} might miss some associations between them. GLRE solved this issue by building the global heterogeneous graph and considering the context relation information, which broke the word sequence.

\item Prior knowledge is required in Case 3. One must know that \textit{``fatigue"} belongs to \textit{``adverse effects"} ahead of time. Then, the relation between \textit{``bepridil"} and \textit{``dizziness"} can be identified correctly. Unfortunately, both GLRE and \citet{wang2019fine} lacked the knowledge, and we leave it as our future work.
\end{compactenum}

\begin{table}
\centering
\resizebox{\columnwidth}{!}{
\begin{tabular}{llll}
	\hline
		\multicolumn{4}{|p{1.12\columnwidth}|}{... [S8] Among \textbf{\textcolor{cyan}{non-users of aspirin}}, the adjusted hazard ratios were: \textbf{\textcolor{red}{rofecoxib}} 1.27, naproxen 1.59, diclofenac 1.17 and ibuprofen 1.05. ... [S10] CONCLUSION: Among \textbf{\textcolor{cyan}{non-users of aspirin}}, naproxen seemed to carry the highest risk for AMI / \textbf{\textcolor{red}{GI bleeding}}. ...} \\
	\hline 
		\textbf{Case 1} & Label: CID & GLRE: CID & \citeauthor{wang2019fine}: N/A \\ \\
	\hline
		\multicolumn{4}{|p{1.12\columnwidth}|}{... [S2] \textbf{\textcolor{cyan}{S-53482}} and \textbf{\textcolor{red}{S-23121}} are N-phenylimide herbicides and produced \textbf{\textcolor{red}{embryolethality}},  teratogenicity. ...} \\
	\hline
		\textbf{Case 2} & Label: CID & GLRE: CID & \citeauthor{wang2019fine}: N/A \\ \\
	\hline
		\multicolumn{4}{|p{1.12\columnwidth}|}{[S1] Clinical evaluation of \textbf{\textcolor{cyan}{adverse effects}} during \textbf{\textcolor{red}{bepridil}} administration for atrial fibrillation and flutter. ... [S8] There was marked QT prolongation greater than 0.55 s in 13 patients ... and general \textbf{\textcolor{red}{fatigue}} in 1 patient each. ...} \\
	\hline
		\textbf{Case 3} & Label: CID & GLRE: N/A & \citeauthor{wang2019fine}: N/A 
\end{tabular}}
\caption{Case study on the CDR test set. CID is short for the \textit{``chemical-induced disease"} relation. \textbf{\textcolor{red}{Target entities}} and \textbf{\textcolor{cyan}{related entities}} are colored accordingly.}
\label{tab:case} 
\end{table}

We analyzed all 132 inter-sentential relation instances in the CDR test set that were incorrectly predicted by GLRE. Four major error types are as follows: (1) Logical reasoning errors, which occurred when GLRE could not correctly identify the relations established indirectly by the bridge entities, account for 40.9\%. (2) Component missing errors, which happened when some component of a sentence (e.g., subject) was missing, account for 28.8\%. In this case, GLRE needed the whole document information to infer the lost component and predict the relation, which was not always accurate. (3) Prior knowledge missing errors account for 13.6\%. (4) Coreference reasoning errors, which were caused by pronouns that could not be understood correctly, account for 12.9\%.

\section{Conclusion}
\label{sect:concl}

In this paper, we proposed GLRE, a global-to-local neural network for document-level RE. Entity global representations model the semantic information of an entire document with R-GCN, and entity local representations aggregate the contextual information of mentions selectively using multi-head attention. Moreover, context relation representations encode the topic information of other relations using self-attention. Our experiments demonstrated the superiority of GLRE over many comparative models, especially the big leads in extracting relations between entities of long distance and with multiple mentions. In future work, we plan to integrate knowledge graphs and explore other document graph modeling ways (e.g., hierarchical graphs) to improve the performance.\\

\noindent\textbf{Acknowledgments.} This work is supported partially by the National Key R\&D Program of China (No. 2018YFB1004300), the National Natural Science Foundation of China (No. 61872172), and the Water Resource Science \& Technology Project of Jiangsu Province (No. 2019046).

\bibliography{emnlp2020}
\bibliographystyle{acl_natbib}


\appendix

\section{Notations}

To help understanding, Table \ref{tab:symbol} summarizes the key notations used in this paper.

\begin{table}[h]
\centering
\resizebox{\columnwidth}{!}{
	\begin{tabular}{c|l}
			\hline \textbf{Symbols} & \textbf{Descriptions} \\ 
			\hline	$\mathcal{D},k$ & a document, the document length \\
					$w,\mathbf{h}$ & a word, the hidden states of a word \\
					$\mathbf{H}$ & the output of BERT \\
					
			\hline	$m,e,s$  & a mention, an entity, a sentence \\
					$\mathbf{n}_{m},\mathbf{t}_m$ & a mention's node rep. \& type rep. \\
					$\mathbf{n}_{e},\mathbf{t}_e$ & an entity's node rep. \& type rep. \\
					$\mathbf{n}_{s},\mathbf{t}_s$ & a sentence's node rep. \& type rep. \\
					
			\hline	$L$ & the number of R-GCN layers \\
					$x,\mathcal{X}$ & an edge type, the set of edge types \\
					$\mathcal{N}$ & a node's neighbors linked by an edge \\
					
			\hline	$\mathcal{Q},\mathcal{K},\mathcal{V}$ & queries, keys, values of multi-head attn. \\
					$z$ & the number of attention heads \\
					$\mathcal{M},\mathcal{S}$ & the sets of mention \& sentence nodes \\
					
			\hline	$\mathbf{e}^\textrm{glo},\mathbf{e}^\textrm{loc}$ & an entity global rep. \& local rep. \\
					$\delta,\mathbf{\Delta}$ & a distance, the distance rep. matrix \\
					$\mathbf{\hat{e}}$ & an entity final rep. \\
					
			\hline	$r,\mathcal{R}$ & a relation, the set of relations \\ 
					$\mathbf{o}_r$ & a target relation rep. \\
					$\mathbf{o}_c$ & a context relation rep. \\
					$\theta$ & the attn. weight for a relation rep. \\
					$\mathbf{y}$ & the probability distribution of a relation \\
					$y^*$ & the true label of a relation \\
			\hline
	\end{tabular}}
\caption{Notations in the paper.}
\label{tab:symbol}
\end{table}

\section{Dataset Availability}

The CDR dataset \cite{li2016biocreative} is available at \url{https://biocreative.bioinformatics.udel.edu/media/store/files/2016/CDR_Data.zip}. The DocRED dataset \cite{yao2019docred} is available at \url{https://github.com/thunlp/DocRED}. Note that, the gold standard of the test set of DocRED is unknown, and only F1 scores can be obtained via an online interface at \url{https://competitions.codalab.org/competitions/20717}.

\section{Experimental Setup}

In this section, we provide more details of our experiments. We implemented GLRE with PyTorch 1.5 and trained it on a server with an Intel Xeon Gold 5117 CPU, 120 GB memory, two NVIDIA Tesla V100 GPU cards and Ubuntu 18.04.

\begin{table}[!b]
\begin{center}
\resizebox{\columnwidth}{!}{
	\begin{tabular}{l|c}
		\hline	Hyperparameters & Values \\ 
		\hline	Batch size & 8 \\
				Learning rate & 0.0005\\
				Gradient clipping & 10 \\
				Early stop patience & 15 \\
				Regularization & $10^{-4}$ \\
				Dropout ratio & 0.2 or 0.5 \\
				
		\hline	Dimension of words & $768$ \\ 
				Dimension of nodes & $256$ \\
				Dimension of node types & $20$ \\
				Number of R-GCN layers & $\textrm{CDR}=3,\textrm{DocRED}=2$ \\
				Number of attention heads & $\textrm{CDR}=4,\textrm{DocRED}=2$ \\
				Dimension of distance & 20 \\
				Final dimension of entities & $532\ (=256\times 2+20)$ \\
				Dimension of relations & $1064\ (=532+532)$ \\
		\hline
	\end{tabular}}
\end{center}
\caption{Hyperparameters in the experiments.}
\label{tab:hyper_param} 
\end{table}

Analogous to \citet{christopoulou2019connecting}, we pre-processed the CDR dataset, including sentence splitting, word tokenization and hypernym filtering. 

When using the training set only, we trained a model on the training set, searched the best epoch in terms of F1 scores on the development set, and tested on the test set. Under the ``train\,+\,dev'' setting, we first trained on the training set and evaluated on the development set, in order to find the best epoch. Then, we re-ran on the union of the training and development sets until the best epoch and evaluated on the test set.
For both cases, we employed dropout and layer normalization \cite{ba2016layer} to prevent model overfitting. 

The parameters of GLRE were initialized with a Gaussian distribution ($\textrm{mean} = 0$ and $\textrm{SD} = 1.0$) using a fixed initialization seed. We trained GLRE by Adam optimizer \citep{kingma2015adam} with mini-batches. The hidden size of BERT was set to 768. A transformation layer was used to project the BERT output into a low-dimensional space of size 256. All hyperparameter values used in the experiments are shown in Table \ref{tab:hyper_param}.

\section{Case Study on DocRED}

In this section, we show a few examples from the DocRED development set in Table \ref{tab:case_docred}, as a supplement to the case study in Section 4.5.
\begin{compactenum}[(1)]
\item Logical reasoning is needed in Case 1. In order to identify the relational fact, one needs to use two mentions of \textit{``Conrad Johnson''} in S1 and S2, respectively. Specifically, one first identifies the fact that \textit{``Conrad Johnson''} was born in \textit{``Texas''} from S2, and then infers the fact that \textit{``Conrad Oberon Johnson''} (coreference) is an \textit{``American''} educator from S1. 

\item Coreference reasoning is needed in Case 2. In order to recognize the relation between \textit{``The Hungry Ghosts''} and \textit{``Michael Imperioli''}, one has to infer that \textit{``He''} refers to \textit{``Michael Imperioli''} in S5.

\item Prior knowledge is needed in Case 3. Through some external prior knowledge, one can know that \textit{``North America''} is a continent and \textit{``California''} is a state, which are the valuable information to help judge their relation. 
\end{compactenum}

We also compare GLRE against the model without entity local representations and the model without context relation representations.
\begin{compactenum}[(4)]
\item In order to predict the relation between \textit{``Kunar''} and \textit{``Afghanistan''}, GLRE attends more to \textit{``Afghanistan''} in [S3] by entity local representations, and correctly identifies the relation. However, GLRE without entity local representations outputs ``N/A''.

\item To predict the relation between \textit{``Breaking Dawn''} and \textit{``Stephenie Meyer''}, GLRE relies on the context relation \textit{``author''} between \textit{``Eclipse''} and \textit{``Stephenie Meyer''}, which is easier to be predicted. In contrast, GLRE without context relation representations imprecisely predicts it as \textit{``creator"} (for general work).
\end{compactenum}

\begin{table}
\centering
\resizebox{\columnwidth}{!}{
\begin{tabular}{llll}
	\hline
		\multicolumn{4}{|p{1.45\columnwidth}|}{[S1] \textbf{\textcolor{cyan}{Conrad Oberon Johnson}} (November 15, 1915--February 3, 2008) was an \textbf{\textcolor{red}{American}} music educator, long associated with the city of Houston, who was inducted into the Texas Bandmasters Hall of Fame in 2000. [S2] Born in Victoria, \textbf{\textcolor{red}{Texas}}, \textbf{\textcolor{cyan}{Conrad Johnson}} was nine when his family moved to Houston. ...} \\
	\hline 
		\textbf{Case 1} & Label: country & GLRE: country & \citeauthor{wang2019fine}: N/A \\ \\
	\hline
		\multicolumn{4}{|p{1.45\columnwidth}|}{[S1] \textbf{\textcolor{red}{Michael Imperioli}} (born March 26, 1966) is an American actor, writer and director best known for ... [S4] He was starring as Detective Louis Fitch in the ABC police drama Detroit 1-8-7 ... [S5] \textbf{\textcolor{cyan}{He}} wrote and directed his first feature film, \textbf{\textcolor{red}{The Hungry Ghosts}}, in 2008. ...} \\
	\hline
		\textbf{Case 2} & Label: director & GLRE: director & \citeauthor{wang2019fine}: cast \\ \\
	\hline
		\multicolumn{4}{|p{1.45\columnwidth}|}{[S1] The Pleistocene coyote (Canis latrans orcutti), also known as the Ice Age coyote, is an extinct subspecies of coyote that lived in western \textbf{\textcolor{red}{North America}} during the Late Pleistocene era. [S2] Most remains of the subspecies were found in southern \textbf{\textcolor{red}{California}}, though at least one was discovered in Idaho. ...} \\
	\hline
		\textbf{Case 3} & Label: continent & GLRE: continent & \citeauthor{wang2019fine}: country \\ \\
	\hline
		\multicolumn{4}{|p{1.45\columnwidth}|}{[S1] Operation Unified Resolve is an air and ground operation to flush out and trap al - Qaeda fighters hiding in the eastern \textbf{\textcolor{red}{Afghanistan}} provinces. [S2] Launched on 23 June 2003, Operation Unified Resolve is a joint operation between Pakistan, United States, and \textbf{\textcolor{red}{Afghanistan}}. [S3] Over 500 troops, mostly from the U.S. 82nd Airborne Division, began hunting the Taliban and al - Qaeda fighters in the provinces of Nangarhar and \textbf{\textcolor{red}{Kunar}} on \textbf{\textcolor{red}{Afghanistan}}’s eastern border. ...} \\
	\hline
		\textbf{Case 4} & Label: country & GLRE: country & w/o local rep.: N/A\\ \\
	\hline
		\multicolumn{4}{|p{1.45\columnwidth}|}{[S1] \textbf{\textcolor{cyan}{Eclipse}} is the third novel in the Twilight Saga by \textbf{\textcolor{red}{Stephenie Meyer}} ... [S3] \textbf{\textcolor{cyan}{Eclipse}} is preceded by New Moon and followed by \textbf{\textcolor{red}{Breaking Dawn}}. ... [S6] Eclipse was the fourth bestselling book of 2008, only behind Twilight, New Moon, and Breaking Dawn. ...} \\
	\hline
		\textbf{Case 5} & Label: author & GLRE: author & w/o context rel.: creator
\end{tabular}}
\caption{Case study on the DocRED development set. \textbf{\textcolor{red}{Target entities}} and \textbf{\textcolor{cyan}{related entities}} are colored.}
\label{tab:case_docred} 
\end{table}

\end{document}